\documentclass[conference]{IEEEtran}
\IEEEoverridecommandlockouts
\usepackage{cite}
\usepackage{amsmath,amssymb,amsfonts}
\usepackage{algorithmic}
\usepackage{subfig}
\usepackage{graphicx}
\usepackage{textcomp}
\usepackage{xcolor}
\usepackage{amsmath}
\usepackage{booktabs}
\usepackage{hyperref}

\newcommand{\pap}{ConCURL }

\def\BibTeX{{\rm B\kern-.05em{\sc i\kern-.025em b}\kern-.08em
    T\kern-.1667em\lower.7ex\hbox{E}\kern-.125emX}}

\begin{document}

\title{Consensus Clustering With Unsupervised Representation Learning
}

\author{\IEEEauthorblockN{ Jayanth Reddy Regatti}
\IEEEauthorblockA{
\textit{The Ohio State University, USA} \\
regatti.1@osu.edu}
\and
\IEEEauthorblockN{ Aniket Anand Deshmukh}
\IEEEauthorblockA{
\textit{Microsoft, USA} \\
aniketde@umich.edu}
\and
\IEEEauthorblockN{ Eren Manavoglu}
\IEEEauthorblockA{
\textit{Microsoft, USA} \\
ermana@microsoft.com}
\and
\IEEEauthorblockN{ Urun Dogan}
\IEEEauthorblockA{
\textit{Microsoft, USA} \\
urundogan@gmail.com}
}

\maketitle

\begin{abstract}
  Recent advances in deep clustering and unsupervised representation learning are based on the idea that different views of an input image (generated through data augmentation techniques) must either be closer in the representation space, or have a similar cluster assignment. Bootstrap Your Own Latent (BYOL) is one such representation learning algorithm that has achieved state-of-the-art results in self-supervised image classification on ImageNet under the linear evaluation protocol. However, the utility of the learnt features of BYOL to perform clustering is not explored. In this work, we study the clustering ability of BYOL and observe that features learnt using BYOL may not be optimal for clustering. We propose a novel consensus clustering based loss function, and train BYOL with the proposed loss in an end-to-end way that improves the clustering ability and outperforms similar clustering based methods on some popular computer vision datasets.  
\end{abstract}


\section{Introduction}

Supervised learning algorithms have shown great progress recently, but generally require a lot of labeled data. However, in many domains (e.g., online advertising, social platforms, etc.), most of the available data are not labeled and manually labeling it is a very labor, time, and cost intensive task \cite{deshmukh2019kernel}. On the other hand, clustering algorithms do not need labeled data to group similar data points into same partitions. Some popular clustering algorithms include k-means, hierarchical clustering, DBSCAN, spectral clustering, etc., and the usefulness of each algorithm varies with the application. These traditional clustering approaches use hand crafted features. However, hand crafted features may not be optimal, and are not scalable to large scale real word datasets \cite{wu2019deep}. 

In this work, we focus on clustering of images along with representation learning. Advancements in deep learning techniques have enabled end-to-end learning of rich representations for supervised learning. For the purposes of clustering, however, such features (learnt via supervised learning) can not be obtained (due to lack of availability of labels), and therefore supervised learning approaches fall short of providing a solution.

Several recent approaches such as SimCLR \cite{chen2020simple},  BYOL~\cite{grill2020bootstrap}, SwAV ~\cite{caron2020unsupervised} have shown great promise in learning end-to-end representations in an unsupervised way. These representation learning algorithms were shown to achieve good results when evaluated using linear evaluation protocol, semi-supervised training on ImageNet, or transfer to downstream tasks \cite{grill2020bootstrap}. A straight forward solution to the clustering problem is to use the features learnt using these algorithms such as BYOL, and apply an out-of-the-box clustering algorithm (such as k-means) to compute clusterings of the data. However, the performance of these features for clustering (using an out-of-the-box clustering algorithm) is not known, and it is possible that these features may not be optimal for clustering. 

On the other hand, simultaneously learning the feature spaces with a clustering objective may lead to degenerate solutions, which until recently limited end-to-end implementations of clustering with representation learning approaches \cite{caron2018deep}. Recent deep clustering works take several approaches to address this issue such as alternating pseudo cluster assignments and pseudo supervised training, comparing the predictions with their own high confidence assignments \cite{caron2018deep,asano2019self,xie2016unsupervised,wu2019deep}, and maximizing mutual information between predictions of positive pairs \cite{ji2019invariant}, etc. Despite the success of these methods, there exists scope for improvement of the clustering performance on several computer vision datasets. We summarize our contributions as follows.
\subsection{Contributions}
\begin{enumerate}
    \item We study the clustering ability of features learnt using BYOL, and observe that these features may not be optimal for clustering for all datasets
    \item To improve clustering performance, we propose to train with a combination of three losses: the BYOL objective, a clustering objective and the proposed consensus clustering objective that can be trained in an end-to-end way. The ensemble in the consensus clustering objective is generated by performing random transformations on the embeddings.
    \item We show empirical results on five popular computer vision datasets based on clustering performance. We show that the proposed algorithm \pap  (Consensus Clustering with Unsupervised Representation Learning) improves upon the clustering ability of BYOL (and also outperforms several other baselines) for three datsets (STL10, CIFAR10, CIFAR100), while BYOL performs strongly on two datasets (ImageNet-10, ImageNet-Dogs).
\end{enumerate}

We now discuss some previous work in clustering and representation learning in more detail.

\subsection{Previous Work}
\paragraph{Clustering: }Clustering is a ubiquitous task and it has been actively used in many different scientific and practical pursuits such as detecting genes from microarray data \cite{frey2007clustering}, segmentation in medical imaging to support diagnosis \cite{masulli1999fuzzy}, etc. We refer interested readers to these excellent sources for a survey of these uses \cite{jain1999data,xu2005survey}. To improve the clustering performance, Strehl and Ghosh~\cite{strehl2002cluster} 
propose a knowledge-reuse framework where multiple clusterings are combined to compute a better clustering. See Ghosh and Acharya~\cite{ghosh2011cluster} for a survey on consensus clustering. 

\paragraph{Self-Supervised Representation Learning: } Self-supervised learning is a sub-field of unsupervised learning in which the main goal is to learn general purpose representations by exploiting user-defined tasks (pretext tasks). There are many different flavors of self-supervised learning such as Instance Recognition tasks \cite{wu2018unsupervised,zhuang2019local}, contrastive techniques \cite{he2020momentum,chen2020simple}. In instance recognition tasks, each image is considered as its own category so that the learnt embeddings are well separated \cite{wu2018unsupervised}. Building on Instance Recognition task, Zhuang \textit{et al.}~\cite{zhuang2019local} propose a Local Aggregation (LA) method based on a robust clustering objective (using multiple runs of k-means) to move statistically similar data points closer in the representation space and dissimilar data points further away.
In contrastive techniques, representations  are  learned  by  maximizing  agreement  between different augmented views of the same data example (known as positive pairs) and minimizing agreement between augmented views of different examples (known as negative pairs). Recent works BYOL \cite{grill2020bootstrap}, SwAV \cite{caron2020unsupervised} achieve state-of-the-art results without requiring negative pairs. We explain these methods in more detail in Section~\ref{sec:method}.

\paragraph{Deep Clustering: } Deep Clustering involves simultaneously learning cluster assignments and features using deep neural networks. DEC \cite{xie2016unsupervised} is one of the first algorithms to show deep learning can be used effectively to cluster images in an unsupervised way, where they use features learnt from an autoencoder to finetune the cluster assignments. DeepCluster \cite{caron2018deep} shows that it is possible to train deep convolutional neural networks end-to-end with pseudo labels that are generated by a clustering algorithm. Subsequently, there have been several works  \cite{ji2019invariant,niu2020gatcluster,wu2019deep,huang2020deep} that introduce end-to-end clustering based objectives and achieve state-of-the-art clustering results. There are other non end-to-end approaches such as SCAN \cite{van2020scan} which uses the learnt representations from a pretext task to find semantically closest images to the given image using nearest neighbors. We describe our proposed end-to-end learning approach in the next section which outperforms similar end-to-end methods.

\section{Clustering and Unsupervised Learning}
\label{sec:method}

In this section, we first motivate the problem of clustering using representations learnt from BYOL, and show that there is scope for improvement in the clustering ability of these features. To address this, the proposed algorithm has three components of the loss function. In Section~\ref{sec:byol} we discuss briefly about the BYOL objective and then introduce the clustering objective to be trained alongside BYOL in an end-to-end way in Section~\ref{sec:soft_clus}. Finally in Section~\ref{sec:concurl} we propose the consensus loss.

Given a set of observations $\mathcal{X} = \{x_i \}_{i=1}^N $, the goal is to simultaneously learn representations and partition $N$ observations into $K$ disjoint clusters {(the value of $K$ is assumed to be known)}. Additionally, we also assume that the data is uniformly distributed in the $K$ clusters which is a common assumption in state of the art works \cite{huang2020deep,niu2020gatcluster}.

As discussed in the previous section, a straightforward approach to perform clustering is to first perform self-supervised representation learning on the data (in this paper, we use BYOL), and then use the learnt representations and apply an out of the box clustering algorithm such as k-means. Several clustering metrics exist to quantify the goodness of clustering, here we use Cluster Accuracy, Normalized Mutual Information (NMI), Adjusted Rand Index (ARI) which are widely used in the literature. 

It is important to note that, unsupervised/self-supervised representation learning algorithms like BYOL are usually evaluated using linear evaluation protocol on ImageNet, or transfer to downstream tasks. 
However, it is not certain if the learnt representations are equally good for clustering.

Table~\ref{tab:comparison_sota} shows the clustering metrics for features  (we used the target projection output which will be introduced in Section~\ref{sec:byol}) trained using BYOL.  It is clear from the table that, BYOL performs poorly on some datasets as compared to the existing state of the art approaches. 
This motivates us to ask the following question.
 \textit{How can we improve the clustering ability of the features learnt using BYOL in an end-to-end and unsupervised way?}

To address the above question, we propose that using a clustering objective along with the BYOL objective while training can improve the clustering ability of features. In particular, we use a soft clustering objective, and propose a novel consensus clustering objective during training and we can observe from Table~\ref{tab:comparison_sota} that it is indeed possible to improve the clustering performance.

We first discuss the BYOL objective now.

\subsection{BYOL objective}
\label{sec:byol}

In BYOL, representations are learnt by enforcing that the latest and bootstrapped embeddings of the augmented views are closer  \cite{grill2020bootstrap}. Bootstrapped embeddings are maintained using a target network that is an exponentially averaged version of the online network (that is trained with gradients). The online network consists of an encoder $f_\theta$ (for example ResNet18), a projector $g_\theta$ such as a 2 layer Multi-Layer Perceptron (MLP), and a predictor $h_\theta$ (e.g. 2 layer MLP) which we explain in the next paragraph. The target network consists of encoder $f_\xi$, and predictor $g_\xi$, where the values of the parameters $\xi$ are computed using an exponential average of $\theta$. Using an MLP to project the representations was shown to be useful in improving the representations of the layer before the MLP \cite{chen2020simple} and thereafter used in Grill \textit{et al.}~\cite{grill2020bootstrap}, Caron \textit{et al.}~\cite{caron2020unsupervised}.

Given an input batch of images $\mathcal{X}_b\subset \mathcal{X}$ (of batch size B), two views $\mathcal{X}_b^1, \mathcal{X}_b^2$ are generated using two randomly chosen image augmentations. Let one input image be $\mathbf{x}\in \mathcal{X}_b$, and $\mathbf{x}^1, \mathbf{x}^2$ be the two views of the input image. For view 1, the online network outputs the representation $y_\theta^1=f_\theta(\mathbf{x}^1)$, projection $v_\theta^1=g_\theta(y_\theta^1)$, and prediction $w_\theta^1=h_\theta(v^1)$, and correspondingly $y_\theta^2, v_\theta^2, w_\theta^2$ for view 2. The target network outputs representation $y_\xi^1=f_\xi(\mathbf{x}^1)$, and projection $v_\xi^1=g_\xi(y_\xi^1)$ for view 1, and similarly $y_\xi^2, v_\xi^2$ for view 2. The projection and prediction outputs are normalized, i.e. $\bar{w}_\theta^1 = \frac{w_\theta^1}{\|w_\theta^1 \|_2}$, $\bar{v}_\xi^2 = \frac{v_\xi^2}{\|v_\xi^2 \|_2}$. The loss is computed by minimizing the mean squared error between the normalized online prediction of view 1 and the normalized target projection of view 2 and vice versa

\begin{align}
\begin{split}
    \mathcal{L}^{BYOL}_{\theta,\xi} &= \big\|{\bar{w}_\theta^1 - \bar{v}_\xi^2 \big\|}_2^2 = 2 - 2\frac{\langle \bar{w}_\theta^1, \bar{v}_\xi^2 \rangle}{ \big\|\bar{w}_\theta^1 \big\|_2 \big\|\bar{v}_\xi^2 \big\|_2}\\
    \tilde{\mathcal{L}}^{BYOL}_{\theta,\xi} &= \big\|{\bar{w}_\theta^2 - \bar{v}_\xi^1 \big\|}_2^2 = 2 - 2\frac{\langle \bar{w}_\theta^2, \bar{v}_\xi^1 \rangle}{ \big\|\bar{w}_\theta^2 \big\|_2 \big\|\bar{v}_\xi^1 \big\|_2}
    \label{eq:cosine-loss}
\end{split}
\end{align}

The parameter weights $\theta$ are updated using gradients of the loss 
\begin{equation}
\label{eq:byol_loss}
    L_1 = \mathcal{L}^{BYOL}_{\theta,\xi} + \tilde{\mathcal{L}}^{BYOL}_{\theta,\xi}
\end{equation}
    
with respect to ${\theta}$ only. The weights $\xi$ are computed using a weighted average i.e., $
\xi \leftarrow \tau_B \xi + (1-\tau_B) \theta 
$, where $\tau_B\in[0,1]$. We use $\tau_B=0.99$ in this paper based on the ablations performed in Grill \textit{et al.}~\cite{grill2020bootstrap}. 

In order to measure the clustering performance of BYOL, we first use k-means algorithm on the features computed in BYOL to compute a cluster assignment. To compute metrics like the clustering accuracy, we are required to solve an assignment problem (computed using a Hungarian match) 
between the true class labels and the cluster assignments, which require equal number of classes. We can use any of the computed vectors $ y_\theta, v_\theta, w_\theta, y_\xi$ or $v_\xi$ to compute the cluster assignment. We observed that the target projection output $v_\xi$ produced the best clustering, and continue to use that throughout the rest of the paper when measuring clustering performance of BYOL. 

From Table~\ref{tab:comparison_sota}, we observe that the features learnt are not optimal for clustering. The objective of BYOL tries to make the representations of different augmentations of the input image similar to each other. This may not be sufficient to move representations of images belonging to the same class closer together. Therefore, we propose to use a clustering objective along with BYOL objective.

\subsection{Soft Clustering}
\label{sec:soft_clus}

In soft clustering, instead of computing a hard cluster assignment of the data into a given number of $K$ clusters (for example, one hot encoding), a soft probability vector of belonging to each cluster is computed \cite{xie2016unsupervised,ji2019invariant}. This makes it easier to define a loss using the soft probabilities and update the parameters using the gradients to enable end-to-end learning. In this section we first introduce a soft clustering objective and then propose the consensus clustering objective.

 In this work, we follow the soft clustering framework presented in Caron \textit{et al.}(SwAV)~\cite{caron2020unsupervised}, which is a centroid based technique and aims to maintain consistency between the clusterings of the augmented views $\mathcal{X}_b^{1}$ and $\mathcal{X}_b^{2}$. We store a set of randomly initialized prototypes $C=\{ \mathbf{c}_1,\cdots,\mathbf{c}_K \} \in \mathbb{R}^{d\times K}$, where $K$ is the number of clusters, and $d$ is the dimension of the prototypes. We use a two layer MLP $g$ (different from $g_\theta, g_\xi$) to project the features $\mathbf{f}^1 = f_\theta(\mathcal{X}_b^1)$ and $\mathbf{f}^2 = f_\theta(\mathcal{X}_b^2)$ to a lower dimensional space (of size $d$). The output of this MLP (referred to as cluster embeddings) is denoted using ${Z}^1 = \{\mathbf{z}_1^1, \ldots, \mathbf{z}_B^1 \}$ and  ${Z}^2  = \{\mathbf{z}_1^2, \ldots, \mathbf{z}_B^2 \}$ (where $B$ is the batch size) for view $1$ and view $2$ respectively. 
 
Soft clustering approaches based on centroids/prototypes often requires one to compute a measure of similarity between the image embeddings and the prototypes \cite{xie2016unsupervised,caron2020unsupervised}. We compute the probability of assigning a cluster $j$ to image $i$ using the normalized vectors $\bar{\mathbf{z}}_i^1 = \frac{\mathbf{z}_i^1}{||\mathbf{z}_i^1||}$, $\bar{\mathbf{z}}_i^2 = \frac{\mathbf{z}_i^2}{||\mathbf{z_i}^2||}$ and $\bar{\mathbf{c}}_j = \frac{\mathbf{c_j}}{||\mathbf{c_j}||}$ as

\begin{align}
\label{eqn:softmaxprob}
\begin{split}
        \mathbf{p}^l_{i,j} &= \frac{\exp (\frac{1}{\tau} \langle \mathbf{\bar{z}}_i^l , \mathbf{\bar{c}}_j \rangle)}{\sum_{j^{'}} \exp (\frac{1}{\tau} \langle \mathbf{\bar{z}}^l_{i} , \mathbf{\bar{c}}_{j^{'}} \rangle)} ,   \quad l\in\{1,2\}
\end{split}
\end{align}

We concisely write $ \mathbf{p}^1_{i} = \{ \mathbf{p}^1_{i,j} \}_{j = 1}^K $ and  $ \mathbf{p}^2_{i} = \{ \mathbf{p}^2_{i,j} \}_{j = 1}^K $. 
Here, $\tau$ is a temperature parameter and we set the value to $0.1$ similar to Caron \textit{et al.}~\cite{caron2020unsupervised}. Note that, we use $\mathbf{p}_{i}$ to denote the predicted cluster assignment probabilities for image $i$ (when not referring to a particular view), and a shorthand $\mathbf{p}$ is used when $i$ is clear from context. 
The idea of predicting assignments $\mathbf{p}$, and then comparing them with high-confidence estimates $\mathbf{q}$ (referred to as codes henceforth) of the predictions was proposed in Xie \textit{et al.}~\cite{xie2016unsupervised}. While Xie \textit{et al.}~\cite{xie2016unsupervised} uses pretrained features (from autoencoder) to compute the predicted assignments and the codes, using their approach in an end-to-end unsupervised manner might lead to degenerate solutions. Asano \textit{et al.}~\cite{asano2019self} avoid such degenerate solutions by enforcing an equi-partition constraint (the prototypes equally partition the data) during code computation using the Sinkhorn-Knopp algorithm \cite{cuturi2013sinkhorn}. Caron \textit{et al.}~\cite{caron2020unsupervised} follow a similar formulation but compute the codes for the two views separately in an online manner for each mini-batch. The assignment codes are computed by solving the following optimization problem

\begin{align}
    \label{eqn:sinkhorn}
    \begin{split}
            Q^i &= \arg\max_{Q\in \mathcal{Q}} \text{Tr}(Q^TC^TZ^i) + \epsilon H(Q),\quad i\in\{1,2\}
    \end{split}
\end{align}

where $ Q = \{\mathbf{q}_1, \ldots, \mathbf{q}_B \} \in \mathbb{R}_{+}^{K\times B}$, $\mathcal{Q}$ is the transportation polytope defined by
\[
\mathcal{Q} = \{ \mathbf{Q}\in \mathbb{R}^{K\times B}_{+}| \mathbf{Q}\mathbf{1}_B = \frac{1}{K}\mathbf{1}_K, \mathbf{Q}^T\mathbf{1}_K = \frac{1}{B}\mathbf{1}_B \}
\]
$\mathbf{1}_K$ is a vector of ones of dimension $K$ and $ H(Q) = -\sum_{i,j}Q_{i,j}\log Q_{i,j}
$. The above optimization is computed using a fast version of the Sinkhorn-Knopp algorithm \cite{cuturi2013sinkhorn} as described in Caron \textit{et al.}~\cite{caron2020unsupervised}.

After computing the codes $Q^1 $ and $Q^2$, inorder to maintain consistency between the clusterings of the augmented views, the loss is computed using the probabilities $\mathbf{p}_{ij}$ and the assigned codes $\mathbf{q}_{ij}$ by comparing the probabilities of view 1 with the assigned codes of view 2 and vice versa, given as

\begin{align}
\label{eqn:swavloss}
    \begin{split}
        L_{2,1} &= - \frac{1}{2B}\sum_{i=1}^{B}\sum_{j=1}^K \mathbf{q}^{2}_{ij} \log \mathbf{p}^{1}_{ij} \\
        L_{2,2} &= - \frac{1}{2B}\sum_{i=1}^{B}\sum_{j=1}^K \mathbf{q}^{1}_{ij} \log  \mathbf{p}^{2}_{ij}, \\
        L_{2} &= L_{2,1} + L_{2,2}
    \end{split}
\end{align}

The parameter weights are trained by using the stochastic gradients of the loss for updates. 

To compute the cluster assignments of the data, it is sufficient to use the computed probability assignments $\{\mathbf{p}_i\}_{i=1}^N$ or the computed codes $\{\mathbf{q}_i\}_{i=1}^N$ (in general we observe that using $\mathbf{q}$ gives better clustering performance and use it throughout the paper) and assign the cluster index as $c_i = \arg \max_{k} \mathbf{q}_{ik}$ for the $i^{\text{th}}$ datapoint. As discussed earlier, to compute metrics like cluster accuracy, we assume that the number of prototypes $K$ is equal to the true number of classes in the dataset. 

To improve the clustering ability of traditional clustering algorithms, Strehl and Ghosh~\cite{strehl2002cluster} propose a knowledge reuse framework by building an unsupervised ensemble using several distinct clusterings of the same data.
Fern and Brodley~\cite{fern2003random} build on the cluster ensemble framework by generating the unsupervised ensemble using multiple random projections of high dimensional data and compute the clusterings of each projection. An aggregated similarity matrix for the clusterings is computed which is used to cluster the data using an agglomerative clustering algorithm. However, in the above framework, the features are fixed. Inspired by this approach, we propose a consensus clustering loss based on the soft clustering objective defined above in order to assist in learning the features and improving clustering performance.

\subsection{Consensus Clustering}
\label{sec:concurl}

We generate a cluster ensemble by first performing transformations on the cluster embeddings $Z^1, Z^2$ and the prototypes $C$. 
It is important to note the distinction between the usage of the terms: views and ensemble in this context. The views are generated from any data augmentation of the image and if both views come from the same image, they should have the closer representations. In this paper, ensemble on the other hand is generated after randomly projecting the cluster embeddings and the prototypes to a new space and then predicting soft cluster assignments in this new space.

{There can be many clusterings for any given latent space (either by randomization introduced in the space or different clustering algorithms). We propose to use consensus clustering so that we can find a latent space that maximizes performance on many different clusterings. To make the method effective, we need more diversity in the clusterings among each component of the ensemble at the start and ideally the diversity should decrease with training and the clustering performance should get better.} We argue that creating such a diverse ensemble leads to better representations and better clusters.

At the beginning of the algorithm, we randomly initialize $M$ such transformations and fix them throughout training. Suppose using a particular random transformation (a randomly generated matrix $A$), we get $\Tilde{\mathbf{z}} = A\mathbf{z},\; \Tilde{\mathbf{c}} = A\mathbf{c}$. We then compute the softmax probabilities $\Tilde{\mathbf{p}}_{ij}$ using the normalized vectors  $\Tilde{\mathbf{z}}/||\Tilde{\mathbf{z}}||$ and $\Tilde{\mathbf{c}}/||\Tilde{\mathbf{c}}||$. Repeating this with the $M$ transformations results in $M$ predicted cluster assignment probabilities for each view. When the network is untrained, the embeddings $\mathbf{z}$ are random and applying the random transformations followed by computing the predicted cluster assignments leads to a diverse set of soft cluster assignments. 

To compute the consensus loss, once the probabilities $\Tilde{\mathbf{p}}_{ij}$ are computed, we compare the codes generated using Eq. (\ref{eqn:sinkhorn}) of view 1 with the $\Tilde{\mathbf{p}}$ of view 2 and vice versa, given as 

\begin{align}
    \begin{split}
        \label{eqn:consloss}
        L_{31} &= -\frac{1}{2BM} \sum_{i=1}^{B}\sum_{m=1}^M \sum_{j=1}^{K}  \mathbf{q}^{2}_{ij} \log \Tilde{\mathbf{p}}^{(1,m)}_{ij} , \\
        L_{32} &= -\frac{1}{2BM} \sum_{i=1}^{B}\sum_{m=1}^M \sum_{j=1}^{K}
        \mathbf{q}^{1}_{ij} \log \Tilde{\mathbf{p}}^{(2,m)}_{ij} \\
        L_3 &= L_{31} + L_{32}
    \end{split}
\end{align}

The final loss that we sought to minimize is the combination of the losses $L_1$ (Eq.~\ref{eq:byol_loss}), $L_2$ (Eq.~\ref{eqn:swavloss}), $L_3 $ (Eq. \ref{eqn:consloss})
\begin{equation}
    \label{eqn:totalloss}
    L_{\text{total}} = \alpha L_1 + \beta L_2 + \gamma L_3.
\end{equation}

{where $\alpha, \beta, \gamma$ are non-negative real valued constants. $L_1$ minimizes the distance between the embeddings of different views of the input batch, $L_2$ maintains consistency between clusterings of two augmented views of the input batch, and $L_3$ maintains consistency between clusterings of the randomly transformed cluster embeddings.} In this paper, we consider $(\alpha, \beta,\gamma)=(1,1,1)$ when discussing about the proposed \pap algorithm, and when $(\alpha, \beta,\gamma)=(1,0,0)$, we refer to clustering with BYOL features.

Similar to Soft Clustering, we compute the cluster assignments using the computed codes $\{\mathbf{q}\}_{i=1}^N$ and compute the cluster metrics.

\section{Understanding the Consensus Objective}

In this section, we will try to build intuition regarding the proposed consensus clustering objective. A key component of consensus clustering is to generate an ensemble of clusterings in an unsupervised way. We discuss some choices of generating the ensemble in Section~\ref{sec:generate_ensemble}, and empirically study the diversity of some such ensembles in Section~\ref{sec:compute_diversity}.

\subsection{Choice of Generating Multiple Clusterings}
\label{sec:generate_ensemble}

\begin{table}[b]
\caption{Different ways to generate ensembles}\label{tab:diffwaysensembles}
{%
\begin{tabular}{p{3.3cm}p{4.4cm}}\\\toprule  
Data Representation & Clustering algorithms \\\midrule
Different data preprocessing techniques & Multiple clustering algorithms (k-means, GMM, etc)\\  \midrule
Subsets of features & Same algorithm with different parameters or initializations\\  \midrule
Different transformations of the features & Combination of multiple clustering and different parameters or initializations\\  \bottomrule
\end{tabular}
}
\end{table}

Fred and Jain,~\cite{fred2005combining} discuss different ways to generate an ensemble of clusterings which are tabulated in Table \ref{tab:diffwaysensembles}. In our proposed algorithm, we focus on the choice of data representation to generate cluster ensembles.

By fixing a stable clustering algorithm, we can generate arbitrarily large ensembles by  applying different transformations on the embeddings. Random projections were successfully used in Consensus Clustering previously \cite{fern2003random}. By generating ensembles using random projections, we have control over the amount of diversity we can induce into the framework, by varying the dimension of the random projection. In addition to Random Projections, we also used diagonal transformations \cite{hsu2018unsupervised} where different components of the representation vector are scaled differently. Hsu \textit{et al.,}~\cite{hsu2018unsupervised} illustrate that such scaling enables a diverse set of clusterings which is helpful for their meta learning task. For the hyper-parameters of the transformations used such as number of transformations, dimension of projection, etc., refer to Section~\ref{sec:expresults}.

\subsection{Computing pairwise NMI of the ensemble}
 \label{sec:compute_diversity}

\begin{figure}[h]
\centering
\subfloat[]{
    \label{fig:NMI_mean}
    \includegraphics[width=0.49\columnwidth]{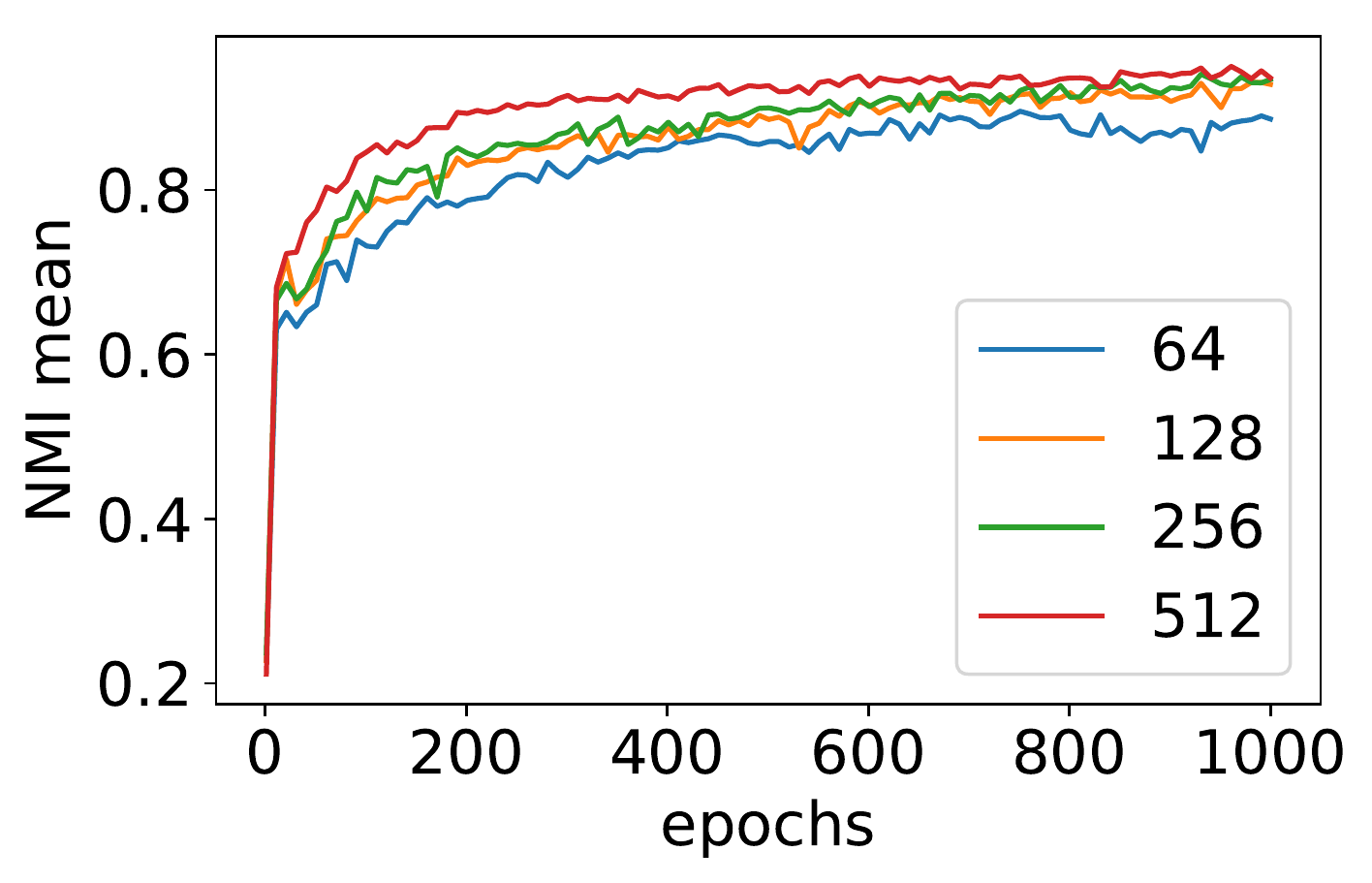}
    } 
\subfloat[]{
    \label{fig:NMI_std}
    \includegraphics[width=0.49\columnwidth]{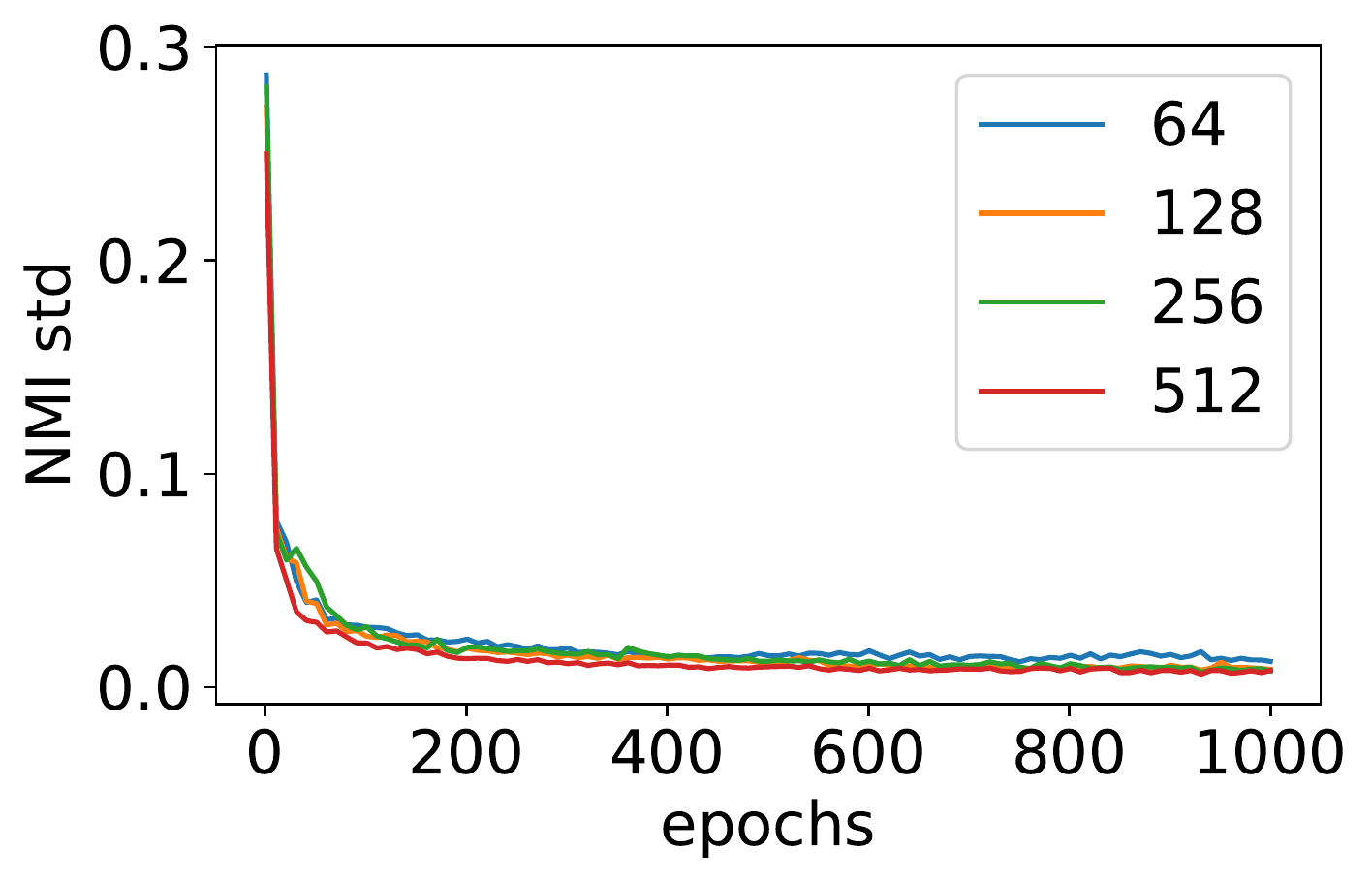}
}
\caption{Pairwise NMI as a way to measure the diversity in the ensemble. The results are for STL10 dataset, and shows pairwise NMI for different random projection dimensions (the original dimension of $\mathbf{z}$ is 256).}
\end{figure}

\begin{table}[htb]
\begin{center}
\caption{Dataset Summary: The resolution column shows the size to which we resized the images in our algorithm.}
  \label{tab:dataset_summary}
\resizebox{\linewidth}{!}{%
\begin{tabular}{||c | c | c | c | c ||} 
 \hline
 Dataset & Classes & Train Data & Test Data & Resolution
 \\ [0.5ex] 
 \hline\hline
 ImageNet-10 
 & 10 & 13000 & 500 & 224$\times$ 224 \\ 
 \hline
 Imagenet-Dogs 
 & 15 & 19500 & 750 & 224$\times$ 224 \\
 \hline
 STL-10 
 & 10 & 5000 & 8000 & 96$\times$ 96 \\
 \hline
 CIFAR10 
 & 10 & 50000 & 10000 & 32$\times$ 32 \\
 \hline
 CIFAR100-20 
 & 20 & 50000 & 10000 & 32$\times$ 32 \\
 \hline
\end{tabular}}
\end{center}
\end{table}

\begin{table*}[thb]
\begin{center}
 \caption{Clustering evaluation metrics: Comparison with other clustering algorithms}
\label{tab:comparison_sota}
\resizebox{\textwidth}{!}{%
\begin{tabular}{|c| ccc | ccc | ccc | ccc | ccc |}
    \hline
     Datasets & 
          \multicolumn{3}{c}{ImageNet-10} & \multicolumn{3}{c}{ImageNet-Dogs} & \multicolumn{3}{c}{STL10} & \multicolumn{3}{c}{CIFAR10} & \multicolumn{3}{c}{CIFAR100-20} \vline\\
    \hline
    Methods\textbackslash Metrics & 
    Acc & NMI & ARI & 
    Acc & NMI & ARI &
    Acc & NMI & ARI & 
    Acc & NMI & ARI &
    Acc & NMI & ARI  \\
    \hline
    DCCM \cite{wu2019deep} &
        0.710 & 0.608 & 0.555 &
        0.383 & 0.321 & 0.182 &
        {0.482} & {0.376} & {0.262} &
        {0.623} & {0.496} & {0.408} &
        {0.327 }& {0.285} & {0.173} \\
    \hline
    GATCluster \cite{niu2020gatcluster} &
        0.762 & 0.609 & 0.572 &
        0.333 & 0.322 & 0.200 &
        {0.583} & {0.446} & {0.363}  &
        {0.610} & {0.475} & {0.402} &
        {0.281} & {0.215} & {0.116} \\
    \hline
    PICA \cite{huang2020deep} &
        0.870 & 0.802 & 0.761 &
        0.352 & 0.352 & 0.201 &
        {0.713} & {0.611} & {0.531} &
        {0.696} & {0.591} & {0.512} &
        {0.337} & {0.310} & {0.171} \\
    \hline    
    BYOL \cite{grill2020bootstrap} &
        \textbf{0.939} & \textbf{0.885} & \textbf{0.868} &
        \textbf{0.542} & \textbf{0.522} & \textbf{0.341} &
        {0.594} & {0.506} & {0.375} &
        0.638 & 0.429 & 0.440 &
        0.313 & 0.287 & 0.153 \\
    \hline
    \pap (Ours) &
        {0.931} & {0.882} & {0.869} &
        {0.447} & {0.45} & {0.298} &
        \textbf{{0.752}} & \textbf{{0.645}} & \textbf{{0.58}} &
        \textbf{{0.785}} & \textbf{{0.667}} & \textbf{{0.614}} &
        \textbf{{0.409}} & \textbf{{0.39}} & \textbf{{0.232}} \\
    \hline

\end{tabular}}
\end{center}
\end{table*}

\begin{table}[b]
\begin{center}
\caption{Clustering evaluation metrics of STL10 by varying $\alpha,\beta,\gamma$}
\label{tab:ablation_losses}
\begin{tabular}{|c| ccc |}
    \hline
     Datasets & 
     \multicolumn{3}{c}{STL10} 
     \vline\\
    \hline
    Methods\textbackslash Metrics & 
    Acc & NMI & ARI 
    \\
    \hline
    BYOL &
        {0.594} & {0.506} & {0.375}
        \\
    \hline
    Soft Clustering &
        0.565 & 0.533 & 0.401 
        \\
    \hline
    BYOL+Soft Clustering &
        0.651 & 0.572 & 0.46 
        \\
    \hline
    ConCURL &
        \textbf{{0.752}} & \textbf{{0.645}} & \textbf{{0.58}}
        \\
    \hline

\end{tabular}%
\end{center}
\end{table}

 We measure the diversity of the ensemble at every epoch to observe the affect of consensus as training progresses. For each component of the ensemble, we use the softmax probability estimates $\tilde{\mathbf{p}}$ and compute cluster assignments by taking an $\arg\max$ on $\tilde{\mathbf{p}}$ of each image. If there are $M$ components in the ensemble, we get $M$ such cluster assignments. We then compute a pairwise NMI (Normalized Mutual Information) (similar analysis to Fern and Brodley,~\cite{fern2003random}) between every two components of the ensemble, and compute the average and standard deviation of the pairwise NMI across the $\frac{M(M-1)}{2}$ pairs. We observe from Figure \ref{fig:NMI_mean} that the pairwise NMI increases as training progresses and becomes closer to 1 as expected. At the beginning of training, the ensemble is very diverse (small NMI score with a larger standard deviation); and as training progresses, the diversity is reduced (large NMI score with a smaller standard deviation).

\section{Experimental Results}
\label{sec:expresults}
We evaluated our algorithm and compared against existing work on some popular image datasets: ImageNet-10, ImageNet-Dogs (subsets of ImageNet \cite{deng2009imagenet}; we used the same classes as \cite{Chang_2017_ICCV}), STL10 \cite{coates2011analysis}, CIFAR-10, CIFAR-100 \cite{krizhevsky2009learning}. For CIFAR-100, we used the 20 meta classes as the class labels while evaluating clustering. The dataset summary is given in Table \ref{tab:dataset_summary}. We evaluate for the cluster accuracy (ACC), Normalized Mutual Information (NMI), Adjusted Rand Index (ARI) of the computed cluster assignments (same metrics used in the state-of-the-art papers \cite{huang2020deep,niu2020gatcluster}). For ImageNet-10 and ImageNet-Dogs, we trained using the train split and evaluated on the train split. For STL10, CIFAR10 and CIFAR-100, similar to earlier approaches (PICA \cite{huang2020deep} and GATCluster \cite{niu2020gatcluster}) we used both train and test splits for training and evaluation unless otherwise specified. Note that PICA also uses unlabelled data split of 100k points in STL10 which we don't use. The open source implementation is publicly available at \href{https://github.com/JayanthRR/ConCURL}{https://github.com/JayanthRR/ConCURL}.

\subsection{Implementation Details}
\label{sec:implementation}

\subsubsection{Architecture}
We use a residual network \cite{he2016deep} with 34 layers (current state of the art clustering results \cite{huang2020deep} also use the same architecture). 
The encoder $f_\theta$ thus outputs a 512 dimension vector (layer before the last fully connected layer) after passing through the network. The projector network $g_\theta$ and predictor network $h_\theta$ in BYOL were chosen similar to Grill \textit{et al.}~\cite{grill2020bootstrap} and have the same architecture; a linear layer with output dimension 4096, followed by batch normalization, and ReLu layers, and a final output layer of dimensions 256. Therefore, the projection and prediction outputs $v_\theta, w_\theta, v_\xi$ are all 256 dimensional. For the clustering part, the MLP projection head $g$ consists of a hidden layer of size 2048 similar to Caron \textit{et al.}~\cite{caron2020unsupervised}, followed by batch normalization and ReLU layers, and an output layer of size 256. The prototypes are thus chosen to be of dimension 256. Please note that the output of the encoder $f_\theta$ is fed as input to both $g_\theta$ and $g$.

We use Adam optimizer with a learning rate of 0.0005 to perform the updates for all datasets. We performed a coarse learning rate search and found 0.0005 to be the best performing value. We implemented our algorithm using the Pytorch framework and trained our algorithm using one V100 GPU. We used a batch size of 256 for CIFAR-10 and CIFAR-100 datasets, and a batchsize of 128 for the remaining datasets. For STL-10 we train for 1000 epochs, and for other datasets, we train for 500 epochs.

\subsubsection{Image Augmentations}
The different views $\mathcal{X}_b^1, \mathcal{X}_b^2$ are not the same as views in multi-view datasets \cite{schops2017multi}. The views referred to in this paper correspond to different augmented views that are generated by image augmentation techniques, such as ‘RandomHorizontalFlip’, ‘RandomCrop’.
In this work, we use the augmentations used in \cite{chen2020simple,grill2020bootstrap}. We first crop a random patch of the image with scale ranging from 0.08 to 1.0, and resize the cropped patch to 224$\times$224 (96$\times$96 in the case of smaller resolution datasets such as STL10). The resulting image was then flipped horizontally with a probability of 0.5. We then apply color transformations, starting by randomly changing the brightness, contrast, saturation and hue with a probability of 0.8. The image was then changed to gray-scale with a probability of 0.2. Then we applied a Gaussian Blur with kernel size 23$\times$23 and a sigma chosen uniformly randomly between 0.1 and 2.0. The probability of applying the Gaussian Blur was 1.0 for view 1 and 0.5 for view 2. During evaluation, we resized the image such that the smaller edge of the image is of size 256 (not required for STL, CIFAR10, CIFAR100-20), and a center crop is performed with the resolution mentioned in Table \ref{tab:dataset_summary}.  
The color transformations were computed using Kornia \cite{eriba2019kornia} which is a differentiable computer vision library for Pytorch.
We finally normalize the image channels with the mean and standard deviation computed on ImageNet. Additionally, we also observed that applying a Sobel filter (after all the image augmentations are performed and before the forward pass) helped in some cases. In such cases, the input channels in the first convolution layer were modified accordingly since applying a Sobel filter reduces the number of channels of the input images to 2.

\subsubsection{Random Transformations}
To compute the random transformations on the embeddings $\mathbf{z}$, we followed two techniques. We used Gaussian random projections with output dimension $d$, and transformed the embeddings $\mathbf{z}$ to the new space with dimension $d$. In Gaussian random projections, the projection matrix is generated by picking rows from a Gaussian distribution such that they are orthogonal. We also used diagonal transformation \cite{hsu2018unsupervised} where we multiply $\mathbf{z}$ with a randomly generated diagonal matrix of the same dimension as $\mathbf{z}$. We initialized $M$ random transformations at the beginning and remain fixed throughout the training.

We performed model selection on the hyperparameters of the random transformations on the embedding space such as the number of random transformations $M$ (10, 30, 50, 100) and the dimensionality of the output space if using a random projection (we used 32, 64, 128, 256, 512).

Note that we fixed the number of prototypes to be equal to the number of ground truth classes. It was shown however that over-clustering leads to better representations \cite{caron2020unsupervised,ji2019invariant,asano2019self} and we can extend our model to include an over-clustering block with a larger set of prototypes \cite{ji2019invariant} and alternate the training procedure between the blocks.

\begin{table*}[htb!]
\caption{Ablations: a) The number of diagonal transformations b) The number of random projections and c) Dimension size of random projections. These results were obtained by training \pap on train+test split of STL10 for 1000 epochs}
\label{tab:Ablation}
\begin{center}
\resizebox{\textwidth}{!}{
\begin{tabular}{ ccc }   
Number of Diagonal Transformations & Number of Random Projections & Embedding Dimension Size \\  
\begin{tabular}{ |c| ccc| } 
    \hline
    \#Trans & Acc & NMI & ARI  \\
    \hline
    10  &
        0.731 & 0.631 & 0.551  \\
    \hline
    30  &
         0.608 & 0.569 & 0.440 \\
    \hline
    50 &
        0.589 & 0.543 &  0.412\\
    \hline
    100 &
        0.511 & 0.476 & 0.316\\ 
    \hline
     \hline
    BYOL &
        {0.594} & {0.506} & {0.375}\\
\hline
\end{tabular} &  
\begin{tabular}{ |c| ccc| } 
    \hline
    \#Proj & Acc & NMI & ARI  \\
    \hline
    10  &
        0.625 & 0.580 & 0.449  \\
    \hline
    30  &
         0.733 & 0.615 & 0.547 \\
    \hline
    50 &
        0.610 & 0.569 &  0.432\\
    \hline
    100 &
        0.752 & 0.645 & 0.581\\ 
    \hline
     \hline
    BYOL &
        {0.594} & {0.506} & {0.375}\\         
\hline
\end{tabular} &  
\begin{tabular}{ |c| ccc| } 
    \hline
    \#dim & Acc & NMI & ARI  \\
    \hline
    32  &
        0.731 & 0.619 & 0.545  \\
    \hline
    64  &
         0.752 & 0.645 & 0.580 \\
    \hline
    128 &
        0.646 & 0.556 &  0.461\\
    \hline
    256 &
        0.705 & 0.586 & 0.515\\ 
     \hline
     \hline
    BYOL &
        {0.594} & {0.506} & {0.375}\\         
\hline
\end{tabular} \\

\end{tabular}
}
\end{center}
\end{table*}

\subsection{Comparison with other algorithms}
In our comparison, we considered some recent state-of-the-art methods that directly solve for a clustering objective in an end-to-end fashion from random initialization, and do not use any prior information (nearest neighbors for example) derived through other pretext tasks. The implementation details of \pap are provided in Section~\ref{sec:implementation}. The results are presented in Table~\ref{tab:comparison_sota}. 

We observe that, in all the datasets considered, \pap outperforms the algorithms considered (DCCM, GATCluster, PICA) on all the three metrics (we provide comparison with Local Aggregation \cite{zhuang2019local} in the Appendix). 
Moreover we also observe improvement over the clustering on features learnt using BYOL in three out of the five datasets considered. On STL10, CIFAR10 and CIFAR100, the proposed algorithm outperforms BYOL by a huge margin on all metrics. Where as on ImageNet-10 and ImageNet-Dogs, BYOL performs better clustering than \pap and all other methods considered here.

Note that ImageNet-10 and ImageNet-Dogs are both subsets of ImageNet for which BYOL was originally designed which explains its superior performance.  However, on other three datasets BYOL performs poorly as compared to the second best algorithm (PICA). By using the proposed loss, we observe huge improvement in the clustering performance for STL10, CIFAR10, CIFAR100.

Note that in this study, we considered $(\alpha, \beta, \gamma)=(1,1,1)$ for ConCURL. This choice of weights ($\alpha, \beta, \gamma$) signifies that each loss $L_1, L_2, L_3$ is present in the objective. Optimizing for the weights is a difficult task and is out of the scope of this work. Instead, to study the effect of the losses individually, we perform an ablation study for STL10 dataset on the weights $\alpha, \beta, \gamma$ of the loss function in Eq.~\ref{eqn:totalloss}. The results are presented in Table~\ref{tab:ablation_losses}. For Soft Clustering, we use $(\alpha, \beta, \gamma)=(0,1,0)$, and for BYOL+Soft Clustering we use $(\alpha, \beta, \gamma)=(1,1,0)$. We observe that, clearly the proposed consensus loss is enabling a better clustering performance as compared to only using the BYOL objective, and/or the Soft Clustering objective.

\subsection{Effect of Number of random transformations and Embedding size}
\label{sec:furtheranalysis}

In order to study the sensitivity of the algorithm to the random transformations, we performed an ablation study for STL10 trained on the train and test split (see Table \ref{tab:dataset_summary}). 

In the first column of the Table \ref{tab:Ablation}, we used diagonal transformations, and varied the number of transformations $M$. For diagonal transformations, the performance worsens as the number of transformations is increased. 
The second column of the Table \ref{tab:Ablation} contains results with a fixed dimension of random projection (=64), and varies the number of transformations $M$. Although the trend is not clear, the best result is observed when using 100 transformations.
The third column of the Table \ref{tab:Ablation} contains results with a fixed number of transformations (=100), and varies the dimension of projection  (the original embedding size is 256). We observe that projecting to 64 dimension gives the best result. The results fluctuate with a margin of $\pm$ 0.07 and still perform comparably to the other baselines in almost all cases.

\begin{figure}
    \centering
    \includegraphics[width=0.8\columnwidth]{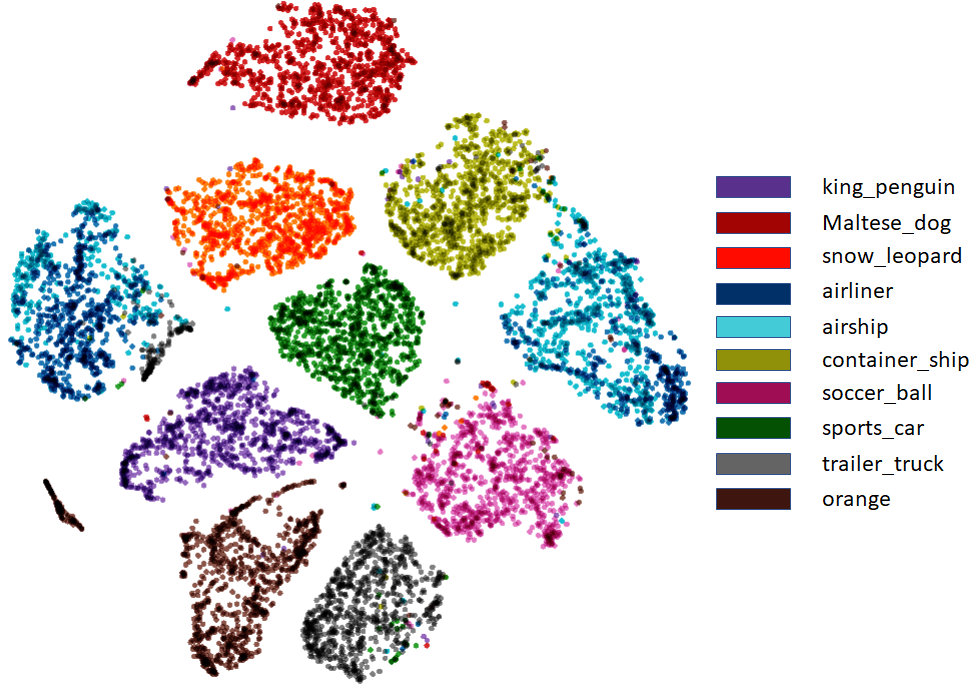}
    \caption{We show t-sne plot of the ImageNet-10 embeddings obtained from \pap trained model. One can clearly see the separation between various clusters with the exception of airline and airship clusters. Airline and airship clusters are mixed together on leftmost and righmost part of the t-sne plot.}
    \label{fig:tsne}
\end{figure}

\subsection{Discussion}
The proposed method achieves good clustering performance on popular computer vision datasets. We can observe the separation between the clusters in the t-SNE plot of the features learnt using \pap in Figure~\ref{fig:tsne}. 

Similar to all the algorithms considered, we assume that ``$K$ - the number of clusters" are known. However, this may not hold true in practice in real world applications. In such a case, we may assume an upper bound on the number of clusters to use as the number of prototypes. Additionally, we also assume that the dataset is equally distributed among the “$K$” clusters. 
If this assumption (also common in literature \cite{huang2020deep,niu2020gatcluster}) does not hold, the fast Sinkhorn-Knopp algorithm used to solve Eq. \ref{eqn:sinkhorn} may not be optimal.

It was observed during the experiments that using only the soft clustering loss $L_2$ to train the network sometimes results in the loss getting stuck at $\ln{K}$ where $K$ is the number of prototypes (we chose equal to true classes) in the data and subsequently leads to degenerate clustering. This issue is also documented in SwaV implementation \footnote{https://github.com/facebookresearch/swav/issues/12}. However, we did not observe this behavior when training with all $L_1, L_2, L_3$ together. We believe one reason for this could be that, adding the other losses act as a regularization and avoid degeneracy in the predicted assignments $\mathbf{p}$ which results in the above behavior. We leave investigating this issue for future work.

\section{Conclusion}
In this work, we motivate the problem of improving the clustering ability of features learnt using BYOL and propose a consensus clustering loss that can be trained with BYOL in an end-to-end way. The proposed method improves the clustering ability of the features, and outperforms several deep clustering algorithms when evaluated on popular computer vision datasets. A possible extension can be to leverage the knowledge reuse framework of \cite{strehl2002cluster} and use the clusterings output by the ensemble to compute a better quality partition of the input data. Also, we believe it would be useful to study if
the proposed approach can be used to improve on the clustering ability of features learnt using other representation learning algorithms. We believe that addressing these questions will further the state-of-the-art in clustering and we consider it as future work.

\bibliographystyle{unsrt}
\bibliography{ijcnn}

\appendix

\section*{Evaluation Metrics}
\label{sec:evalmetrics}

\paragraph{Cluster Accuracy} The clustering accuracy is computed by first computing a cluster partition of the input data. Once the partitions are computed and cluster indices assigned to each input data point, the linear assignment map is computed using Kuhn-Munkres (Hungarian) algorithm that reassigns the cluster indices to the true labels of the data. Clustering accuracy is then given by
$
ACC = \frac{ \sum_{i = 1}^N \mathbb{I}\{y_{true}(x_i) = c(x_i)\}}{N},
$
where $y_{true}(x_i) $ is a true label of $x_i$ and $c(x_i)$ is the cluster assignment produced by an algorithm (after Hungarian mapping). 
\paragraph{Normalized Mutual Information} For two clusterings $U,V$, with each containing $|U|, |V|$ clusters respectively, and let $|U_i|$ be the number of samples in cluster $U_i$ of clustering $U$ (similarly for $V$) , Mutual Information (MI) is given by
$
MI(U,V) = \sum_{i=1}^{|U|}\sum_{j=1}^{|V|} \frac{|U_i \cap V_i|}{N}\log \frac{N|U_i\cap V_j|}{|U_i||V_j|}
$
where $N$ is the number of data points under consideration. Normalized Mutual Information is defined as
$
NMI(U,V) = \frac{MI(U,V)}{\sqrt{MI(U,U)MI(V,V)}}
$

\paragraph{Adjusted Rand Index}\footnote{https://scikit-learn.org/stable/modules/clustering.html\#adjusted-rand-score} 
Suppose R is the groundtruth clustering and S is a partition, the RI of S is given as follows. Let $a$ be the number of pairs of elements that are in the same set in R as well as in S; $b$ be the number of pairs of elements that are in diferent sets in R, and different sets in S. Then $RI = \frac{a + b}{{n \choose 2}},$ and $ ARI = \frac{RI - \mathbb{E}[RI] }{\max(RI) -  \mathbb{E}[RI] }$.

\section*{Comparison with Local Aggregation}

Local Aggregation (LA) \cite{zhuang2019local} builds on Non-parametric Instance Discrimination \cite{wu2018unsupervised} and uses a robust clustering objective (uses a closest neighbors set generated using multiple runs of k-means) similar to consensus clustering to move statistically similar points closer in the representation space and dissimilar points farther away. They demonstrate that the learnt representations are better by evaluating using the linear evaluation protocol on ImageNet. 

However, the performance of these features for clustering was not discussed. Since LA is in spirit similar to our work, we perform a study on the clustering performance of features learnt using LA \footnote{using official Pytorch implementation available at https://github.com/neuroailab/LocalAggregation-Pytorch/tree/master/config} on ImageNet-10, ImageNet-Dogs datasets. The results are presented in Tables~\ref{tab:comparison_la_im10},~\ref{tab:comparison_la_imdogs}.

\begin{table}[htb!]
\begin{center}
\caption{Comparison with LA on ImageNet-10 } 
\label{tab:comparison_la_im10}
\begin{tabular}{|c| ccc |}
    \hline
     Datasets & 
     \multicolumn{3}{c}{ImageNet-10} 
     \vline\\
    \hline
    Method\textbackslash Metrics & 
    Acc & NMI & ARI 
    \\
    \hline
    LA ResNet18 &
        0.336 & 0.282 & 0.161 
        \\
    \hline
    LA ResNet34 &
        0.393 & 0.346 & 0.213 
        \\
    \hline
    ConCURL-ResNet18 &
        0.869 & 0.822 &  0.767 
        \\
    \hline
    ConCURL-ResNet34 &
        {0.931} & {0.882} & {0.869}
        \\
    \hline

\end{tabular}%
\end{center}
\end{table}

\begin{table}[htb!]
\begin{center}
\caption{Comparison with LA on ImageNet-Dogs } 
\label{tab:comparison_la_imdogs}
\begin{tabular}{|c| ccc |}
    \hline
     Datasets & 
     \multicolumn{3}{c}{ImageNet-Dogs} 
     \vline\\
    \hline
    Method \textbackslash Metrics & 
    Acc & NMI & ARI 
    \\
    \hline
    LA ResNet18 &
        0.201 & 0.157 & 0.063  
        \\
    \hline
    LA ResNet34 &
        0.201 & 0.133 & 0.06
        \\
    \hline
    ConCURL-ResNet18 &
        0.356 & 0.377 & 0.207
        \\
    \hline
    ConCURL-ResNet34 &
        {0.447} & {0.45} & {0.298}
        \\
    \hline

\end{tabular}%
\end{center}
\end{table}

We observe that the clustering performance of our proposed algorithm ConCURL is much better than the clustering performance of LA method. Note that clustering performance of BYOL features is better than LA and our algorithm further improves on the clustering performance of BYOL. One major difference between our work and LA is the way in which we generate the ensemble. Our approach allows us to control and measure the diversity of the ensemble which can be useful in making algorithm design choices. Although LA controls the ensemble by varying
(number of clusterings) and
(number of clusters in each clustering) which aptly suits the objective they are solving, their ensembles are limited to use k-means clustering (the authors show that other clustering approaches are either not scalable, or are not optimal). In our case, by applying feature space transformations, we have much more freedom in generating the ensemble. We used random projections, diagonal transformations, but there could be other transformations on the feature space that we have not explored yet.

\subsection{Implementation Details of LA}
We trained the model for 500 epochs. Since the original implementation was for ImageNet, we performed a hyper-parameter search as follows. We took the config file from the official PyTorch implementation and created 36 configuration files by varying learning rate and kmeans ‘k’. In the original config file of authors, they used kmeans-k = 30000 (for 1.28 million images). We scale the kmeans-k for ImageNet-10 (13000 images) and ImageNet-Dogs (19500 images) accordingly and try six different values. In particular, we try learning-rates = [0.003, 0.005, 0.01, 0.03, 0.05, 0.1], and kmeans-k = [10, 15, 100, 310, 452, 500].
 
The background neighbors are 4096 as used in the paper. We ran ResNet18 experiments for the full hyperparameter search (36 experiments) and evaluated the cluster metrics. Additionally, we chose the top 5 choices of hyperparameters from above, and also ran ResNet34 experiments with those parameters, for both the datasets.

Since we report ConCURL results on ResNet34, we additionally trained ConCURL on ResNet18 for a fair comparison with LA results (these are not necessarily the best results with ResNet18 since we did not perform a hyper-parameter search for ConCURL)

The code repository uses a different version of ResNet (PreActResNet). Therefore, for evaluation of cluster performance, we take the output of the layer before the final dense layer. For ResNet18, the output dimensions of this layer is (512,7,7) and we take the mean along (1,2) dimensions and use the resulting 512 dimension vector. We compute k-means clustering on these embeddings using Faiss for the train split of the data, and compute the cluster metrics as mentioned in our paper.

\end{document}